\documentclass{article}

\usepackage{microtype}
\usepackage{graphicx}
\usepackage{subfigure}
\usepackage{booktabs} %

\usepackage{bm}
\usepackage{MyMathCmds}
\usepackage{todonotes}
\usepackage{amsthm}

\newtheorem{theorem}{Theorem}

\usepackage{pdfpages}
\usepackage{filecontents}

\usepackage{hyperref}
\usepackage{xr-hyper}

\usepackage{cleveref}

\usepackage[accepted]{icml2020}

\icmltitlerunning{Sparse Gaussian Processes with Spherical Harmonic Features}

\makeatletter
\newcommand*{\addFileDependency}[1]{%
  \typeout{(#1)}
  \@addtofilelist{#1}
  \IfFileExists{#1}{}{\typeout{No file #1.}}
}
\makeatother

\newcommand*{\myexternaldocument}[1]{%
    \externaldocument{#1}%
    \addFileDependency{#1.tex}%
    \addFileDependency{#1.aux}%
}

\myexternaldocument{suppl}

\newcommand{\ourmethod}{{\sc{VISH}}\xspace}

\newcommand{\rkhs}{\ensuremath{{\mathcal{H}}}}
\newcommand{\sphere}{\ensuremath{\mathbb{S}}}
\newcommand{\dsphere}{\ensuremath{\sphere^{d-1}}}

\newcommand{\dnumharmonicsforlevel}{\ensuremath{N^d_\ell}\xspace}
\newcommand{\sh}{\ensuremath{\phi}}

\newcommand{\mat}[1]{\bm{\mathrm{#1}}} %
\renewcommand{\vec}[1]{\bm{\mathrm{#1}}}

\newcommand{\MX}{\mat{X}}

\newcommand{\MZ}{\mat{Z}}

\newcommand{\MS}{\mat{S}}

\newcommand{\MK}{\mat{K}}

\newcommand{\Kuu}{\MK_{\vu \vu}}

\newcommand{\Kff}{\MK_{\vf \vf}}
\newcommand{\kf}{\vec{k}_{\vf}}
\newcommand{\kfx}{\vec{k}_{\vf}(\vx)}

\newcommand{\ku}{\vec{k}_\vu}

\begin{document}

\twocolumn[
\icmltitle{Sparse Gaussian Processes with Spherical Harmonic Features} 

\begin{icmlauthorlist}
\icmlauthor{Vincent Dutordoir}{pio}
\icmlauthor{Nicolas Durrande}{pio}
\icmlauthor{James Hensman}{amazon}
\end{icmlauthorlist}

\icmlaffiliation{pio}{PROWLER.io, Cambridge, United Kingdom}
\icmlaffiliation{amazon}{Amazon Research, Cambridge, United Kingdom (work done while JH was affiliated to PROWLER.io)}

\icmlcorrespondingauthor{Vincent Dutordoir}{vincent@prowler.io}

\icmlkeywords{Gaussian processes, spherical harmonics, variational inference, probabilistic modelling}

\vskip 0.3in
]

\printAffiliationsAndNotice{}  %

\begin{abstract}
We introduce a new class of inter-domain variational Gaussian processes (GP) where data is mapped onto the unit hypersphere in order to use spherical harmonic representations. Our inference scheme is comparable to variational Fourier features, but it does not suffer from the curse of dimensionality, and leads to diagonal covariance matrices between inducing variables. This enables a speed-up in inference, because it bypasses the need to invert large covariance matrices. Our experiments show that our model is able to fit a regression model for a dataset with 6 million entries two orders of magnitude faster compared to standard sparse GPs, while retaining state of the art accuracy. We also demonstrate competitive performance on classification with non-conjugate likelihoods.
\end{abstract}

\section{Introduction}
\label{intro}

Gaussian processes (GPs) \citep{rasmussen2006} provide a flexible framework for modelling unknown functions: they are robust to overfitting, offer good predictive uncertainty estimates and allow us to incorporate prior assumptions into the model. Given a dataset with some inputs $\MX \in \Reals^{N \times d}$ and outputs $\vy \in \Reals^N$, a GP regression model assumes $y_i = f(\vx_i)+\varepsilon_i$ where $f$ is a GP over $\Reals^d$ and where the $\varepsilon_i$ are normal random variables accounting for observation noise. The model predictions at $\vx \in \Reals^d$ are then given by the posterior distribution $f \given \vy$. However, computing the posterior distribution usually scales $\BigO(N^3)$, because it requires solving a system involving an $N \times N$ matrix.

A range of sparse GP methods (see \citet{quinonero2005unifying} for an overview) have been developed to improve on this $\BigO(N^3)$ scaling. Among them, variational inference is a practical approach allowing regression \citep{titsias2009}, classification \citep{hensman2015scalable}, minibatching \citep{hensman2013} and structured models including latent variables, time-series and depth \citep{titsias2010bayesian, frigola2014variational, hensman2014nested, salimbeni2017doubly}. Variational inference in GPs works by approximating the exact (but intractable) posterior $f \given \vy$. This approximate posterior process is constructed from a conditional distribution based on $M$ pseudo observations of $u_i = f(\vz_i)$ at locations $\MZ =  \{\vz_i\}_{i=1}^M$. The approximation is optimised by minimising the Kullback-Leibler divergence between the approximate posterior and the exact one.
The resulting complexity is $\BigO(M^3 + M^2N)$, so choosing $M \ll N$ enables significant speedups compared to vanilla GP models.
However, when using common stationary kernels such as Mat\'ern, Squared Exponential (SE) or rational quadratic, the influence of pseudo-observations are only local and limited to the neighbourhoods of the inducing points $\MZ$, so a large number of inducing points $M$ may be required to cover the input space. This is especially problematic for higher dimensional data as a result of the curse of dimensionality. %
\begin{figure}[t!]
    \centering
    \includegraphics[trim=2.2cm 1.8cm 1.1cm 2.3cm, clip, width=6cm]{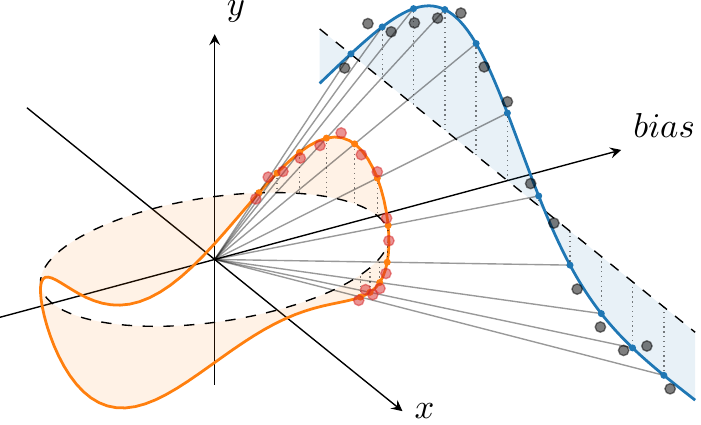}
    \caption{Illustration of the mapping between a 2D dataset (grey dots) embedded into a 3D space and its projection (orange dots) onto the unit half-circle using a linear mapping.}
    \vspace{-.2cm}
    \label{fig:mapping}
\end{figure}

Variational Fourier Features~\citep[VFF]{hensman2017variational} inducing variables have been proposed to overcome this limitation. In VFF the inducing variables based on pseudo observations are replaced with inducing variables obtained by projecting the GP onto the Fourier basis. This results in inducing variables that have global influence on the predictions.
For one-dimensional inputs, this construction leads to covariances matrices between inducing variables that are almost diagonal and this can be exploited to reduce the complexity to $\BigO(N + M^2 N)$.
However, for $d$-dimensional input spaces VFF requires construction of a new basis given by the outer product of the one-dimensional Fourier basis. This implies that the number of inducing variables grows exponentially with the dimensionality, which limits the use of VFF to just one dimension or two. Furthermore, VFF is restricted to kernels of the Mat\'ern family.

In this work we improve upon VFF in multiple directions. Rather than using a sine and cosine basis, we use a basis of spherical harmonics to define a new interdomain sparse approximation. As we will show, spherical harmonics are the eigenfunctions of stationary kernels on the hypersphere, which allows us to exploit the Mercer representation of the kernel for defining the inducing variables. In arbitrary dimensions, our method leads to \emph{diagonal} covariance matrices which makes it faster than VFF as we fully bypass the need to compute expensive matrix inverses. Compared to both sparse GPs and VFF, our approximation scheme suffers less from the curse of dimensionality. As VFF, each spherical harmonic inducing function has a global influence, but there is a natural ordering of the spherical harmonics that can guarantee that the best features are picked given an overall budget for the number $M$ of inducing variables. Moreover, our method works for any stationary kernel on the sphere.

\begin{figure}[t]
    \centering
    \includegraphics[width=\columnwidth]{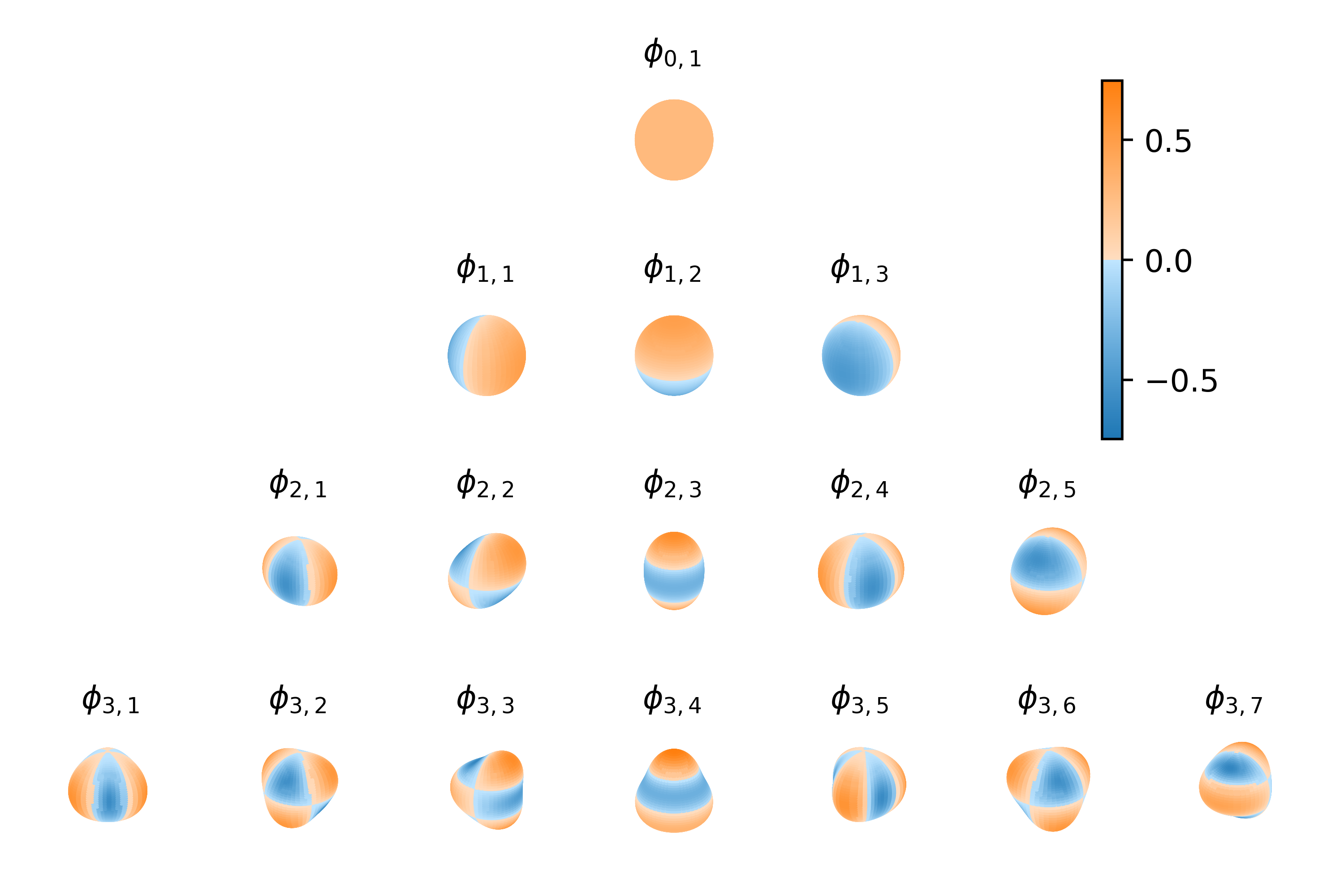}
    \vspace{-.5cm}
    \caption{The first four levels of spherical harmonic functions in $\Reals^3$. The domain of the spherical harmonics is the surface of the unit sphere $\sphere^2$. The function value is given by the color and radius.}
    \label{fig:harmonics}
\end{figure}

Following the illustration in \cref{fig:mapping}, we outline the different steps of our method. We start by concatenating the data with a constant input (bias) and project it linearly onto the unit hypersphere $\dsphere$. We then learn a sparse GP on the sphere based on the projected data. We can do this extremely efficiently by making use of our spherical harmonic inducing variables, shown in \cref{fig:harmonics}. Finally, the linear mapping between the sphere and the sub-space containing the data can then be used to map the predictions of the GP on the sphere back to the original $(\vx, y)$ space.

This paper is organised as follows. In \cref{sec:background} we give the necessary background on sparse GPs and VFF. 
In \cref{sec:vish} we highlight that every stage of the proposed method can be elegantly justified. For example, the linear mapping between the data and the sphere is a property of some specific covariance functions such as the arc-cosine kernel, from which we can expect good generalisation properties (\cref{sec:mapping}). Another example is that the spherical harmonics are coupled with the structure of the Reproducing Kernel Hilbert Space (RKHS) norm for stationary kernels on the sphere, which makes them very natural candidates for basis functions for sparse GPs (\cref{sec:rkhs}). In \cref{sec:definition-sh-inducing-variables} we define our spherical harmonic inducing variables and compute the necessary components to do efficient GP modelling. \Cref{sec:experiments} is dedicated to the experimental evaluation.

\section{Background}
\label{sec:background}

\subsection{GPs and Sparse Variational Inference}

GPs are stochastic processes such that the distribution of any finite dimensional marginal is multivariate normal. The distribution of a GP is fully determined by its mean $\mu(\cdot)$ and covariance function (kernel) $k(\cdot, \cdot)$. GP models typically consist of combining a latent (unobserved) GP $f\sim \GP(0, k(\cdot, \cdot))$ with a likelihood that factorises over observation points $p(\vy \given f) = \prod_n p(y_n \given f(\vx_n))$. When the observation model is $y_n \given f(\vx_n) \sim \NormDist{f(\vx_n), \tau^2}$, the posterior distributions of $f$ and $y$ given some data are still Gaussian and can be computed analytically. Let $\varepsilon_i \sim \NormDist{0, \tau^2}$ represent the observation noise at $\vx_i$, then the GP posterior distribution is $f \given \vy \sim \GP(m, v)$ with
\begin{align*}
    m(\vx) &=  \kfx (\Kff + \tau^2 I)\inv \vy,\\
    v(\vx, \vx') &= k(\vx, \vx') - \kfx\transpose (\Kff + \tau^2 I)\inv \kf (\vx'),
\end{align*}
where $\Kff =\left[ k(\vx_i, \vx_j)\right]_{i,j = 1}^N$ and $\kfx = [k(\vx_n, \vx)]_{n=1}^N$.

Computing this exact posterior requires inverting the $N \times N$ matrix $\Kff$, which has a $\BigO(N^3)$ computational complexity and a $\BigO(N^2)$ memory footprint. Given a typical current hardware specification, this limits the dataset size to the order of few thousand observations. Furthermore, there is no known analytical expression for posterior distribution when the likelihood is not conjugate, as encountered in classification for instance.

Sparse GPs combined with variational inference provide an elegant way to address these two shortcomings \citep{titsias2009, hensman2013, hensman2015scalable}.
It consists of introducing a distribution $q(f)$ that depends on some parameters, and finding the values of these parameters such that $q(f)$ gives the best possible approximation of the exact posterior $p(f \given \vy )$. Sparse GPs introduce $M$ pseudo inputs $\MZ \in \Reals^{M \times d}$ corresponding to $M$ inducing variables $\vu \sim \NormDist{\vm, \MS}$, and choose to write the approximating distribution as $q(f) = q(\vu)p(f\given f(\MZ) \shorteq \vu)$. This results in a distribution $q(f)$ that is parametrised by the variational parameters $\vm \in \Reals^M$, and $\MS \in \Reals^{M\times M}$, which are learned by minimising the Kullback–Leibler (KL) divergence $\KL{q(f)}{p(f \given \vy )}$.

At prediction time, the conjugacy between $q(\vu)$ and the conditioned posterior $f \given f(\MZ) \shorteq \vu$ implies that $q(f)$, where $\vu$ is marginalised out, is a GP with a mean $\mu(\vx)$ and a covariance function $ \nu(\vx, \vx')$ that can be computed in closed:
    \begin{align}
   \mu(\vx) &= \ku(\vx) \Kuu^{-1} \vm     \label{eq:qf}
 \\
  \nu(\vx, \vx') &= k(\vx, \vx') + \ku(\vx)\transpose \Kuu^{-1}(\MS - \Kuu) \Kuu^{-1} \ku(\vx'), \nonumber
    \end{align}
where $\left[ \Kuu \right]_{m,m'} = \ExpSymb[f(\vz_m)\, f(\vz_{m'})]$, and $\left[ \ku(\vx) \right]_{m} = \ExpSymb[f(\vz_m)\,f(\vx)]$.

Sparse GPs result in $\BigO(M^2 N + M^3)$ and $\BigO(M^3)$ computational complexity at training (minimising the KL) and prediction respectively. Picking $M \ll N$ can thus result in drastic improvement, but the lower $M$ is, the less accurate the approximation, as recently shown by \citet{shi2019sparse}. Typical kernels---such as Mat\'ern or Squared Exponential (SE)---depend on a lengthscale parameter that controls how quickly the correlation between two evaluations of $f$ drops for two inputs that move away from another. For short lengthscales, this correlation drops quickly and two observations can be almost independent even for two input points that are close by in the input space. For a sparse GP model, this implies that $\mu(\vx)$ and $\nu(\vx, \vx)$ will rapidly revert to the prior mean and variance when $\vx$ is not in the immediate neighbourhood of an inducing point $\vz_i$. A similar effect can be observed when the input space is high-dimensional: because inducing variables only have a local influence, the number of inducing points required to cover the space grows exponentially with the input space dimensionality. In practice, it may thus be required to pick large values for $M$ to obtain accurate approximate posteriors but this defeats the original intent of sparse methods. 

This behaviour where the vector of features $\ku (\cdot)$ of the approximate distribution are given by kernel function $k(\vz_i, \cdot)$ can be addressed using interdomain inducing variables.

\subsection{Inter-domain GPs and Variational Fourier Features}
Inter-domain Gaussian processes (see \citet{lazaro2009inter} and \citet{GPflow2020multioutput} for a modern exposition) use alternative forms of inducing variables such that the resulting sparse GP models result in more informative features.
In interdomain GPs the inducing variables are obtained by integrating the GP $f$ with an inducing function:
\begin{equation*}
    u_m = \int f(\vx)\,g_m(\vx)\,\calcd{\vx}\,.
\end{equation*}
This redefinition of $\vu$ implies that the expressions of $\Kuu$ and $\ku$ change, but the inference scheme of interdomain GPs and the mathematical expressions for the posterior mean and variance are exactly the same as classic sparse GPs.
Depending on the choice of $g_m(\cdot)$, interdomain GPs can result in various feature vector $\ku(\cdot)$. These feature vectors can alleviate the classic sparse GP limitation of inducing variables having only a local influence.%

VFF~\citep{hensman2017variational} is an interdomain method where the inducing variables are given by a Mat\'ern RKHS inner product between the GP and elements of the Fourier basis:
\begin{equation*}
u_m = \langle f, \psi_m \rangle_\rkhs,
\end{equation*}
where $\psi_0 = 1$, $\psi_{2i}=\cos(i x)$ and $\psi_{2i+1}=\sin(i x)$ if the input space is $[0, 2 \pi]$. This leads to
\begin{equation*}
    \Kuu = \left[\langle \psi_i , \psi_{j} \rangle_\rkhs^{} \right]_{i, j = 0}^{M-1} \text{ and } \ku (x) = \left[ \psi_i(x)\right]_{i = 0}^{M-1}\, .
\end{equation*}
This results in several advantages. First, the features $\ku(x)$ are exactly the elements of the Fourier basis, which are independent of the kernel parameters and can be precomputed. Second, the matrix $\Kuu$ is the sum of a diagonal matrix plus low rank matrices. This structure can be used to drastically reduce the computational complexity, and the experiments showed one or two orders of magnitude speed-ups compared to classic sparse GPs. Finally, the variance of the inducing variables typically decays quickly with increasing frequencies, which means that by selecting the first $M$ elements of the Fourier basis we pick the features that carry the most signal.
\begin{figure}[t!]
    \centering
    \vspace{.3cm}
    \includegraphics[trim=0 1.2cm 0 1.25cm, height=3.5cm]{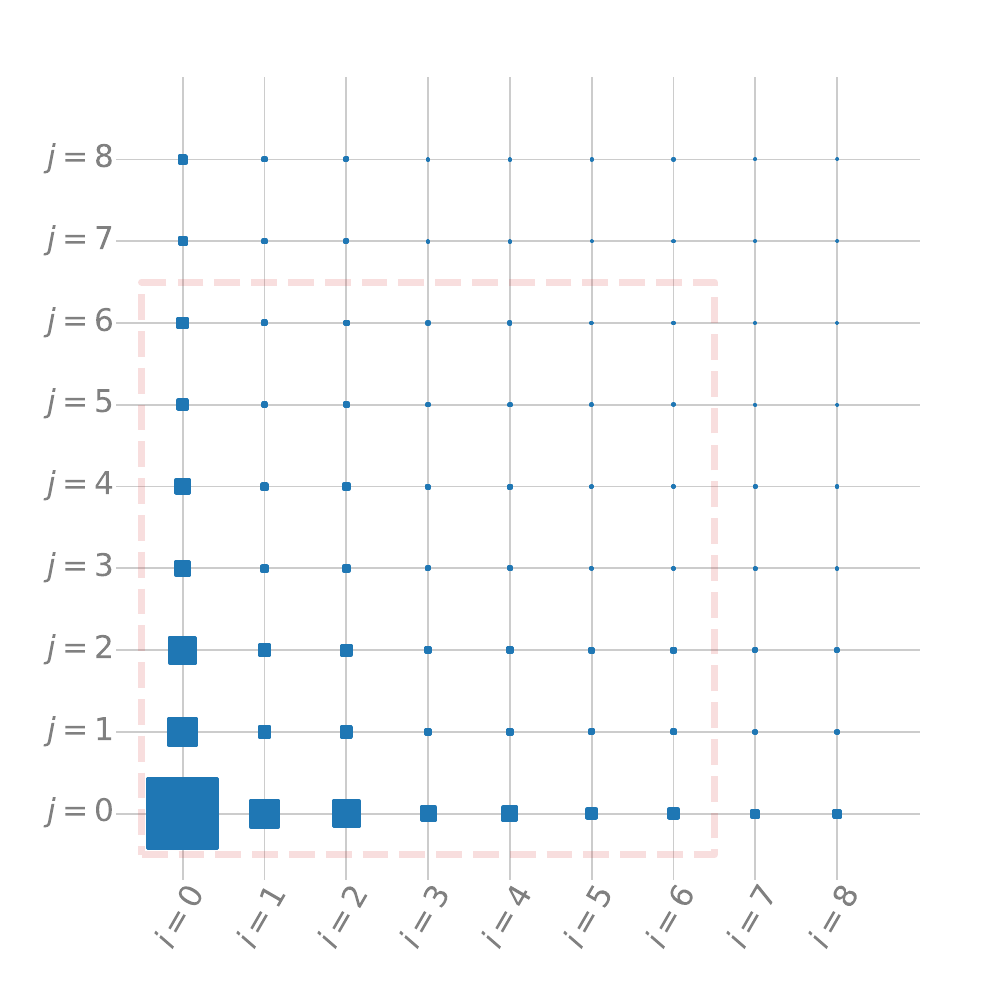}
    \caption{Illustration of the variance of the inducing variables when using VFF with a two dimensional input space. Each pair $(i, j)$ corresponds to an inducing function $\psi_i(x_1)\psi_j(x_2)$, and the area of each square is proportional to the variance of the associated inducing variable. For a given number of inducing variables, say $M=49$ as highlighted with the red dashed line, VFF selects features that do not carry signal (north-east quarter of the red dashed square) whereas it ignores features that are expected to carry signal (along the axes $i=0$ and $j=0$).}
    \label{fig:variance_decay_VFF}
\end{figure}

The main flaw of VFF comes from the way it generalises to multidimensional input spaces. The approach in~\citet{hensman2017variational} for getting a set of $d$-dimensional inducing functions consists of taking the outer product of $d$ univariate basis and to consider separable kernels so that the elements of $\Kuu$ are given by the product of the inner products in each dimension. For example, in dimension 2, a set of $M^2$ inducing functions is given by $\{(x_1, x_2) \mapsto \psi_i(x_1)\psi_j(x_2)\}_{0 \leq i, j \leq M-1}$, and entries on $\Kuu$ are $\langle \psi_i\psi_j, \psi_k\psi_l \rangle_{\mathcal{H}_{k_1 k_2}^{}}^{} = \langle \psi_i, \psi_k \rangle_{\mathcal{H}_{k_1}^{}}^{} \langle \psi_j, \psi_l \rangle_{\mathcal{H}_{k_2}^{}}^{}$. This construction scales poorly with the dimension: for example choosing a univariate basis as simple as $\{1, \cos, \sin\}$ for an eight-dimensional problem already results in more that 6,500 inducing functions. Additionally, this construction is very inefficient in terms of captured variance, as we illustrate in  \Cref{fig:variance_decay_VFF} for a 2D input space. The figure shows that the prior variance associated with the inducing function $\psi_i(x_1)\psi_j(x_2)$ vanishes quickly when both $i$ and $j$ increase. This means that most of the inducing functions on which the variational posterior is built are irrelevant, whereas some important ones such as $\psi_i(x_1)\psi_0(x_2)$ or $\psi_0(x_1)\psi_j(x_2)$ for $i, j \geq \sqrt{M}$ are important but ignored. Although we used a 2D example to illustrate this poor behaviour, it is important to bear in mind that the issue gets exacerbated for higher dimensional input spaces. As detailed in \cref{fig:variance_decay_SH} and discussed later, our proposed approach does not suffer from such behaviour.

\section{Variational Inference with Spherical Harmonics}
\label{sec:vish}
In this section we describe the three key ingredients of the proposed approach: the mapping of dataset to the hypersphere, the definition of GPs on this sphere, and the use of spherical harmonics as inducing functions for such GPs.

\subsection{Linear Mapping to the Hypersphere}
\label{sec:mapping}

As illustrated in \cref{fig:mapping}, for a 1D dataset, the first step of our approach involves appending the dataset with a dummy input value $b$ that we will refer to as the \emph{bias} (grey dots in \cref{fig:mapping}). %
For convenience we will overload the notation, and refer to the augmented inputs $(\vx, b)$ as $\vx$ from now on, denote its dimension by $d$, and assume $x_d=b$. %
The next step is to project the augmented data onto the unit hypersphere $\dsphere$ according to a linear mapping: $(\vx, y) \mapsto (\vx/\norm{\vx}, y/\norm{\vx})$ as depicted by the orange dots. Given the projected data, we learn a GP model on the sphere (orange function), and the model predictions can be mapped back to the hyperplane (blue curve) using the inverse linear mapping (i.e. following the fine grey lines starting from the origin).

Although this setup may seem very arbitrary, it is inspired by the important works on the limits of neural networks as Gaussian processes, especially the \emph{arc-sine} kernel \citep{williams1998computation} and the \emph{arc-cosine} kernels \citep{cho2009kernel}. An arc-cosine kernel corresponds to the infinitely-wide limit of a single-layer ReLU-activated network with Gaussian weights. Let $\vx, \vx' \in \Reals^{d}$, such that $x_d = x'_d = b$, be an augmented input vector whose last entry corresponds to the bias. Then the arc-cosine kernel can be written as
\begin{multline*}
    k_{ac}(\vx, \vx') = \frac{1}{\pi} \norm{\vx}\,\norm{\vx'}\,\big( \sin \theta + (\pi - \theta) \cos \theta \big), \\
    \text{where}\quad\theta = \arccos\left(\frac{\vx\transpose\vx'}{\norm{\vx}\norm{\vx'}}\right).
\end{multline*}
$\theta$ is the geodesic distance (or the great-circle distance) between the projection of $\vx$ and $\vx'$ on the unit hypersphere.
We observe that $k_{ac}$ is separable in polar coordinates, and the dependence in the radius is just the linear kernel. Since the linear kernel is degenerate, it means that there is a one-to-one mapping between the GP samples restricted to the unit sphere $\dsphere$ and the samples over $\Reals^d$, and that this mapping is exactly the linear transformation we introduced previously. %

One insight provided by the link between our settings and the arc-cosine kernel is that once mapped back to the original space, our GP samples will exhibit linear asymptotic behaviour, which can be compared to the way ReLU-activated neural networks operate. Although a proper demonstration would require further work, this correspondence suggests that our models may inherit the desirable generalisation properties of neural networks.

Having mapped the data to the sphere, we must now introduce GPs on the sphere and their covariance. The following section provides some theory that allows us to work with various kernels on the sphere, including the arc-cosine kernel as a special case.

\subsection{Mercer's Theorem for Zonal Kernels on the Sphere}
\label{sec:rkhs}

Stationary kernels, i.e. translation invariant covariances, are ubiquitous in machine learning when the input space is Euclidean. When working on the hypersphere, their spherical counterpart are zonal kernels, which are invariant to rotations. More precisely, a kernel $k: \dsphere \times \dsphere \mapsto \Reals$ is called zonal if there exists a shape function $s$ such that $k(\vx, \vx') = s(\vx\transpose\vx')$. From hereon, we focus on zonal kernels. Since stationary kernels are functions of $\vx - \vx'$ they have the property that $\Delta_\vx k(\vx, \vx') = \Delta_{\vx'}  k(\vx, \vx')$, where $\Delta_\vx = \sum_{i=1}^d \partial^2 / \partial^2 x_i$ is the Laplacian operator. Such property is also verified by zonal kernels: $\Delta^{\dsphere}_\vx k(\vx, \vx') = \Delta^{\dsphere}_{\vx'} k(\vx, \vx')$, where we denote by $\Delta_\vx^{\dsphere}$ the Laplace-Beltrami operator with respect to the variable $\vx$. Combined with an intregration by part, this property can be used to show that the kernel operator $\mathcal{K}$ of a zonal covariance and the Laplace-Beltrami operator commute:
\begin{align*}
    \mathcal{K} \left[\Delta^{\dsphere} g\right] &= \int_{\dsphere} k(\vx, \cdot) \left[\Delta^{\dsphere}_\vx g(\vx)\right] \calcd{\vx}\\
    &= \int_{\dsphere} g(\vx) \Delta^{\dsphere}_\vx k(\vx, \cdot)  \calcd{\vx}\\
    &= \int_{\dsphere} g(\vx) \Delta^{\dsphere} k(\vx, \cdot)  \calcd{\vx} = \Delta^{\dsphere} \mathcal{K} g
\end{align*}
which in turn implies that these two operators share the same eigenfunctions. This result is of particular relevance to us since there is a huge body of literature on diagonalisation of the Laplace-Beltrami operator on $\dsphere$, and that it is well known that its eigenfunctions are given by the spherical harmonics. This reasoning can be summarised by the following theorem:
\begin{theorem}[Mercer representation]
Any zonal kernel $k$ on the hypersphere can be decomposed as
\begin{equation}
\label{eq:kernel-form}
    k(\vx, \vx') = \sum_{\ell=0}^{\infty} \sum_{k=1}^{\dnumharmonicsforlevel} \widehat{a}_{\ell, k} \sh_{\ell, k}(\vx) \sh_{\ell, k}(\vx'),
\end{equation}
where $\vx,\vx' \in \dsphere$ and $\widehat{a}_{\ell, k}$ are positive coefficients, $\sh_{\ell,k}$ denote the elements of the spherical harmonic basis in $\dsphere$, and $N_\ell^d$ corresponds to the number of spherical harmonics for a given level $\ell$.
\end{theorem}
Although it is typically stated without a proof, this theorem is already known in some communities (see \citet{wendland2005} for a functional analysis exposition, or \citet{peacock1999cosmological} for its use in cosmology).

Given the Mercer representation of a zonal kernel, its RKHS can be characterised by
\begin{equation*}
    \rkhs = \left\{
    g = 
    \sum_{\ell=0}^\infty \sum_{k=1}^{\dnumharmonicsforlevel} \widehat{g}_{\ell,k} \sh_{\ell, k} :
    \sum_{\ell=0}^\infty \sum_{k=1}^{\dnumharmonicsforlevel} \frac{|\widehat{g}_{\ell,k}|^2}{\widehat{a}_{\ell, k}} < \infty
    \right\}
\end{equation*}
with the inner product between two functions $g(\vx) = \sum_{\ell, k} \widehat{g}_{\ell, k} \sh_{\ell, k}(\vx)$ and $h(\vx) = \sum_{\ell, k} \widehat{h}_{\ell, k} \sh_{\ell, k}(\vx)$ defined as
\begin{equation}
    \langle g, h \rangle_{\rkhs} = 
        \sum_{\ell=0}^\infty 
        \sum_{k=1}^{\dnumharmonicsforlevel}
            \frac{\widehat{g}_{\ell, k} \widehat{h}_{\ell, k}}%
                 {\widehat{a}_{\ell, k}}.
\end{equation}
It is straightforward to show that this inner product satisfies the reproducing property (see Appendix \cref{appendix:proof:reproducing}).

In order to make these results practical, we need to compute the coefficients $\widehat{a}_{\ell, k}$ for some kernels of interest. For a given value of $\vx' \in \dsphere$ we can see $\vx \mapsto k(\vx, \vx')$ as a function from $\dsphere$ to $\Reals$, and represent it in the basis of spherical harmonics:
\begin{equation}
\label{eq:kernel-funk}
    k(\vx, \vx') = \sum_{\ell=0}^\infty \sum_{k=1}^{\dnumharmonicsforlevel} \langle k(\vx', \cdot), \sh_{\ell, k} \rangle_{L^2(\dsphere)} \sh_{\ell, k}(\vx).
\end{equation}
Combining equations (\ref{eq:kernel-form}) and (\ref{eq:kernel-funk}) gives the following expression for the coefficients we are interested in: $\widehat{a}_{\ell, k} = \langle k(\vx', \cdot), \sh_{\ell, k} \rangle_{L^2(\dsphere)}/ \sh_{\ell, k}(\vx')$.
Although this expression involves a $(d-1)$ dimensional integral on the hypersphere, our hypothesis that $k$ is zonal means we can make use of the Funk-Hecke formula (see \cref{appendix:theorem:funk} in the supplementary) and rewrite it as a simpler 1D integral over the shape function of $k$. Following this procedure finally leads to
\begin{equation}
\label{eq:eigenvalues}
    \widehat{a}_{\ell, k}
    = \frac{\omega_d}{C_{\ell}^{(\alpha)}(1)} \int_{-1}^1\!\!s(t)\,C_\ell^{(\alpha)}(t) (1 - t^2)^{\frac{d-3}{2}} \calcd{t},
\end{equation}
where $\alpha = \frac{d-2}{2}$, $C_\ell^{(\alpha)}$ is the Gegenbauer polynomial of degree $\ell$ and $s(\cdot)$ is the kernel's shape function. The constants $\omega_d$ and $C_{\ell}^{(\alpha)}(1)$ are given in \cref{appendix:theorem:funk} in the supplementary. %

Using \cref{eq:eigenvalues} we are able to compute the Fourier coefficients of any zonal kernel. The details of the calculations for the arc-cosine kernel restricted to the sphere is given in \cref{sec:appendix:arc-cosine} in the supplementary material. 

Alternatively, a key result from \citet[eq. 20]{solin2014hilbert} is to show that the coefficients $\widehat{a}_{\ell, k}$ have a simple expression that depends on the kernel spectral density $S$ and the eigenvalues of the Laplace-Beltrami operator. For GPs on $\dsphere$, the coefficients boil down to $\widehat{a}_{\ell, k } =  S(\sqrt{\ell (\ell + d - 2)})$. This is the expression we used in our experiments to define Mat\'ern and SE covariances on the hypersphere. More details on this method are given in \cref{appendix:sec:materns} in the supplementary.

As one may have noticed, the values of $\widehat{a}_{l,k}$ do not depend on the second index $k$ (i.e. the eigenvalues only depend on the degree of the spherical harmonic, but not on its orientation). This is a remarkable property of zonal kernels which allows us to use the \emph{addition theorem} (see supplementary material \cref{appendix:theorem:addition}) for spherical harmonics to simplify \cref{eq:kernel-form}:
\begin{equation*}
    k(\vx, \vx') = \sum_{\ell=0}^{\infty}  \widehat{a}_{\ell}\, \frac{\ell + \alpha}{\alpha}\, C_\ell^{(\alpha)}\left(\vx\transpose\vx'\right).
\end{equation*}
This representation is cheaper to evaluate than \cref{eq:kernel-form} but it still requires truncation for practical use.

\subsection{Spherical Harmonics as Inducing Features}
\label{sec:definition-sh-inducing-variables}

We can now build on the results from the previous section to propose powerful and efficient sparse GPs models. We want features $\ku(\vx)$ that exhibit non-local influence for expressiveness, and inducing variables that induce sparse structure in $\Kuu$ for efficiency. We achieve this by defining the inducing variables $u_m$ to be the inner product between the GP\footnote{Although $f$ does not belong to $\rkhs$ \citep{kanagawa2018gaussian}, such expression is well defined since the regularity of $\sh_{m}$ can be used to extend the domain of definition of the first argument of the inner product to a larger class of functions. See \citet{hensman2017variational} for a detailed discussion.} and spherical harmonics:\footnote{Note that in the context of inducing variables, we switch to a single integer $m$ to index the spherical harmonics and order them first by increasing level $\ell$, and then by increasing $k$ within a level.}
\begin{equation}
  u_m = \langle f, \sh_m\rangle_{\rkhs}.
\end{equation}

To leverage these new inducing variables we need to compute two quantities: 1) the covariance between $u_m$ and $f$ for $\ku (\vx)$, and 2) the covariance between the inducing variables themselves for the $\Kuu$ matrix. See \citet{GPflow2020multioutput} for an in-depth discussion of these concepts.

The covariance of the inducing variables with $f$ are
\begin{equation*}
   \left[\ku (\vx)\right]_m = \ExpSymb[f(\vx)\,u_m] %
    = \langle k(\vx, \cdot), \sh_m \rangle_{\rkhs} 
    = \sh_{m}(\vx),
\end{equation*}
where we relied on the linearity of both expectation and inner product and where we used the reproducing property of $\rkhs$.

The covariance between the inducing variables is given by
\begin{equation*}
    \left[\Kuu \right]_{m, m'} = \ExpSymb \left[u_m\,u_{m'} \right] 
    = \langle \sh_{m}, \sh_{m'} \rangle_{\rkhs} 
    = \frac{\delta_{mm'}}{\widehat{a}_m}\, ,
\end{equation*}
where $\delta_{mm'}$ is the Kronecker delta. Crucially, this means that $\Kuu$ is a diagonal matrix with elements $1/(\widehat{a}_m)$.

Substituting $\Kuu$ and $\ku (\vx)$ into the sparse variational approximation (\cref{eq:qf}), leads to the following form for $q(f)$
\begin{equation*}
    \GP\left(\tilde{\bm{\Phi}}^\top(\vx) \vm;\ k(\vx, \vx') - \tilde{\bm{\Phi}}^\top(\vx) (\MS - \Kuu) \tilde{\bm{\Phi}}(\vx') \right),
\end{equation*}
with $\tilde{\bm{\Phi}}(\vx) = [\widehat{a}_m \phi_m(\vx)]_{m=1}^M$.

This sparse approximation has two main differences compared to a SVGP model with standard inducing points. First, the spherical harmonic inducing variables lead to features $k_\vu(\vx)$ with non-local structure. Second, the approximation $q(f)$ does not require any inverses anymore. The computational bottleneck of this model is now simply the matrix multiplication in the variance calculation, which has a $\BigO(N_{\text{batchsize}} M^2)$ cost. Compared to the $\BigO(M^3 + N_{\text{batchsize}} M^2)$ cost of inducing point SVGPs, this gives a significant speedup -- as we show in the experiments.

\begin{figure}[t!]
    \centering
    \vspace{.3cm}
    \includegraphics[trim=0 1.5cm 0 1cm, height=3.5cm]{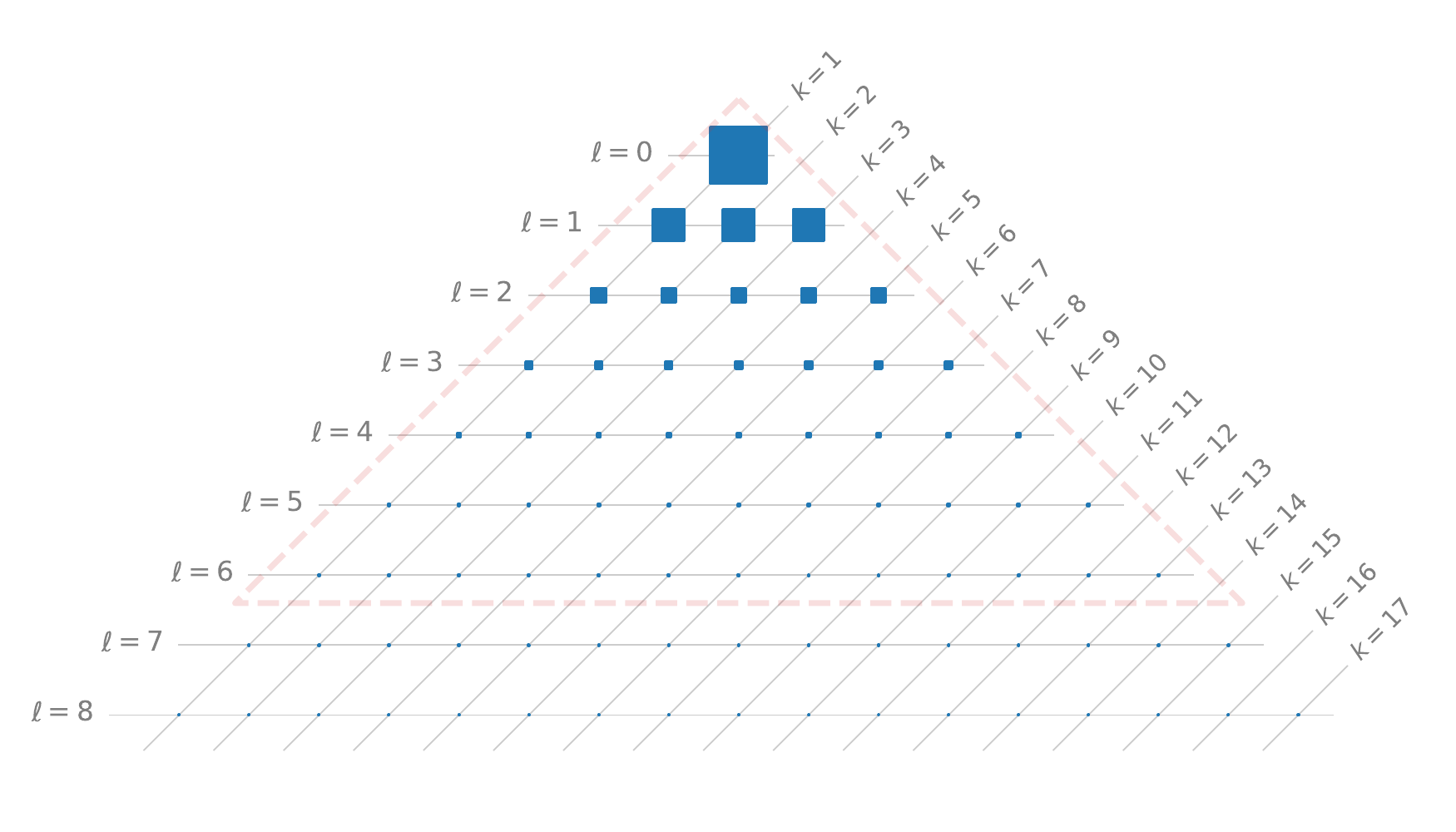}
    \caption{Illustration of the variance of the \ourmethod inducing variables for a 2D input space. Settings are the same as in \cref{fig:variance_decay_VFF} but a spherical harmonic feature $\sh_{\ell, k}$ is associated to each pair $(\ell, k)$. For a given number $M$ of inducing variables, the truncation pattern (see red dashed triangle for $M=49$) is optimal since it selects the most influential features.}%
    \label{fig:variance_decay_SH}
\end{figure}

As is usual in sparse GP methods, the number of inducing variables $M$ is constrained by the computational budget available to the user. Given that we ordered the spherical harmonic by increasing $\ell$, choosing the first $M$ elements means we will select first features with low angular frequency. Provided that the kernel spectral density is a decreasing function (this will be true for classic covariances, but not for quasi-periodic ones), this means that the selected features correspond to the ones carrying the most signal according to the prior. In other words, the decomposition of the kernel can be compared to an infinite dimensional principal component analysis, and our choice of the inducing function is optimal since we pick the ones with the largest variance. This is illustrated in \cref{fig:variance_decay_SH}, which shows the analogue of \cref{fig:variance_decay_VFF} for spherical harmonic inducing functions.

\section{Experiments}
\label{sec:experiments}

We evaluate our method Variational Inference with Spherical Harmonics (\ourmethod) on regression and classification problems and show the following properties of our method: 1) \ourmethod performs competitively in terms of accuracy and uncertainty quantification on a range of problems from the UCI dataset repository. 2) \ourmethod is extremely fast and accurate on large-scale conjugate problems (approximately 6 million 8D entries in less than 2 minutes). 3) Compared to VFF, \ourmethod can be applied to multi-dimensional datasets and preserve its computational efficiency. 4) On problems with non-conjugate likelihood our method does not suffer from some of the issues encountered by VFF.

We begin with a toy experiment in which we show that the approximation becomes more accurate as we increase the number of basis functions.

\begin{figure*}[t]
    \centering
    \includegraphics[width=.9\textwidth]{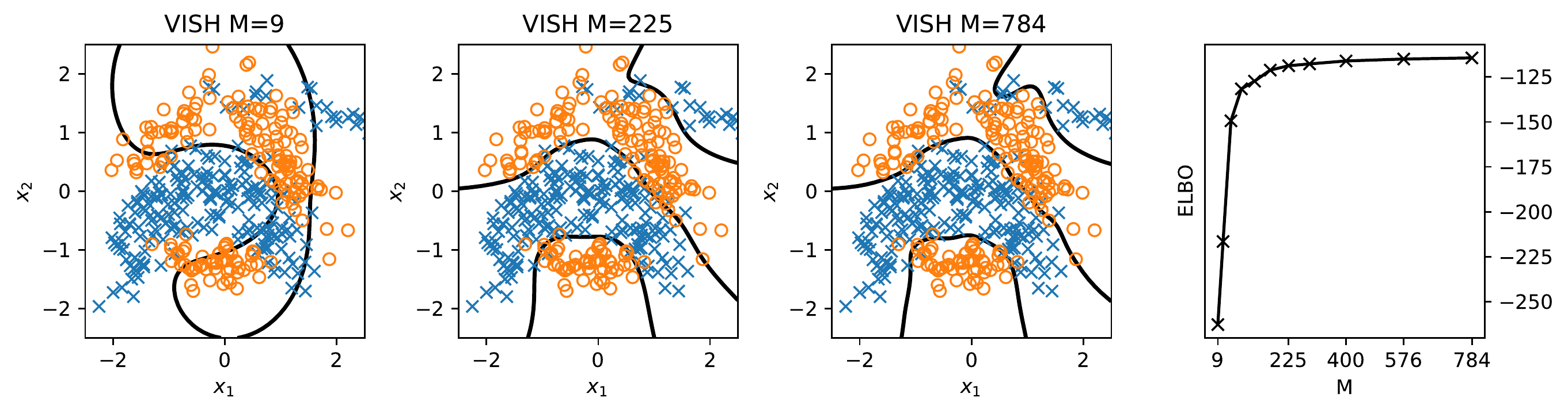}
    \vspace{-.5cm}
    \caption{Classification of the 2D banana dataset with growing number of spherical harmonic basis functions. The right plot shows the convergence of the ELBO with respect to increasing numbers of basis functions.}
    \label{fig:banana}
\end{figure*}

\subsection{Toy Experiment: Banana Classification}

The banana dataset is a 2D binary classification problem \citep{hensman2015scalable}. In \cref{fig:banana} we show three different fits of \ourmethod with $M\in \{9,\ 225,\ 784\}$ spherical harmonics, which correspond respectively to maximum levels of $2$, $14$, and $27$ for our inducing functions. Since the variational framework provides a guarantee that more inducing variables must be monotonically better \citep{titsias2009}, we expect that increasing the number of inducing functions will provide improved approximations. This is indeed the case as we show in the rightmost panel: with increasing $M$ the ELBO converges and the fit becomes tighter.

While this is expected behaviour for SVGP methods, it is not guaranteed by VFF. Given the Kronecker-structure used by VFF for this 2D experiment \citet{hensman2017variational} report that using a full rank covariance matrix for the variational distribution was intolerably slow. They also show that enforcing the Kronecker structure on the posterior results in an ELBO that {\em decreased} as frequencies were added, and they finally propose a sum-of-two-Kroneckers structure, but provide no guarantee that this would converge to the exact posterior in the limit of larger $M$. In \ourmethod we do not need to impose any structure on the approximate covariance matrix $\MS$, so we retain the guarantee that adding more basis functions will move us closer to the posterior process. The method remains fast despite optimising over full covariance matrices: fitting the models displayed in \cref{fig:banana} only takes a few seconds on a standard desktop.

\subsection{Regression on UCI Benchmarks}

\begin{table*}[tb]
\centering
\resizebox{.95\textwidth}{!}{\begin{tabular}{@{}lccccccccccccc@{}}
\toprule
&&&& \multicolumn{5}{c}{MSE}                     
& \multicolumn{5}{c}{NLPD}  \\ 
\cmidrule(lr){5-9} \cmidrule(lr){10-14}
Dataset  &
N$\downarrow$ &
D &
M &
\ourmethod &
A-VFF &
SVGP & 
A-GPR & 
GPR & 
\ourmethod & 
A-VFF & 
SVGP & 
A-GPR & 
GPR \\ \midrule

Yacht    &
$308$  &
$6$ &
$294$ &
{$0.004 \pm 0.00$} &
$0.010 \pm 0.01$ &
$0.002 \pm 0.00$ &
$0.010 \pm 0.00$ &
\bm{$0.001 \pm 0.00$} &
{$-1.698 \pm 0.21$} &
$-0.861 \pm 0.36$ &
$-1.72 \pm 0.12$ &
$-0.903 \pm 0.16$ &
\bm{$-2.420 \pm 0.15$} \\
 
Energy   &
$768$  &
$8$ &
$210$ &
\bm{$0.003 \pm 0.00$}  &
$0.011 \pm 0.00$ &
$0.013 \pm 0.00$ &
$0.012 \pm 0.00$ &
\bm{$0.003 \pm 0.00$} &
\bm{$-1.575 \pm 0.13$} &
$-0.824 \pm 0.11$ &
$-0.61 \pm 0.04$ &
$-0.797 \pm 0.11$ &
$-1.461 \pm 0.12$ \\

Concrete &
$1030$ & 
$8$ &
$210$ &
{$0.122 \pm 0.02$}  &
$0.123 \pm 0.01$ &
$0.128 \pm 0.01$ &
$0.099 \pm 0.01$ &
\bm{$0.096 \pm 0.01$} & 
{$0.336 \pm 0.07$}  &
$0.371  \pm 0.06$ &
$0.374 \pm 0.03$ &
$0.268 \pm 0.06$ &
\bm{$0.228 \pm 0.12$} \\

Kin8mn   &
$8192$ &
$8$ &
$210$ &
{$0.219 \pm 0.01$}  &
$0.561 \pm 0.02$ &
$0.116 \pm 0.01$ &
$0.556\pm 0.02$ &
\bm{$0.076 + 0.00$} & 
{$0.612 \pm 0.02$}  &
$1.129  \pm 0.01$ &
$0.381 \pm 0.03$ &
$1.125 \pm 0.02$ &
\bm{$0.116 \pm 0.02$} \\

Power    &
$9568$ &
$4$ &
$336$ &
{$0.054 \pm 0.00$}  &
$0.058 \pm 0.00$ &
$0.055 \pm 0.00$ &
$0.032 \pm 0.00$ &
\bm{$0.046 \pm  0.00$} &
{$-0.005 \pm 0.03$} &
$-0.002 \pm 0.04$ &
$-0.029 \pm 0.03$ &
$-0.306 \pm 0.07$ &
\bm{$-0.114 \pm 0.04$} \\ \bottomrule

\end{tabular} }
\caption{Predictive mean squared errors (MSEs) and negative log predictive densities (NLPDs) with one standard deviation based on 5 splits on 5 UCI regression datasets. Lower is better. All models assume a Gaussian noise model, use a Mat\'ern-3/2 kernel and use the L-BFGS optimiser for the hyper- and variational parameters. \ourmethod and SVGP are configured with the same number of inducing points $M$. A-VFF and A-GPR assume an Additive structure over the inputs (see text). For A-VFF and \ourmethod the optimal posterior distribution for the inducing variables is set following \citet{titsias2009}.}
\label{tab:uci}
\end{table*}

We use five UCI regression datasets to compare the performance of our method against other GP approaches. We measure accuracy of the predictive mean with Mean Squared Error (MSE) and uncertainty quantification with mean Negative Log Predictive Density (NLPD). For each dataset we randomly select 90\% of the data for training and 10\% for testing and repeat this 5 times to get error bars. We normalise the inputs and the targets to be centered unit Gaussian. We report the MSE and NLPD of the normalised data. In \Cref{tab:uci} we report the performance of \ourmethod, Additive-VFF (A-VFF) \citep{hensman2017variational}, SVGP \citep{hensman2013}, Additive-GRP (A-GPR) and GPR. 

We start by comparing \ourmethod against SVGP, and notice that for Energy, Concrete and Power the change in inductive bias and expressiveness of spherical harmonic inducing variables opposed to standard inducing points improves performance. Also, while the sparse methods (A-VFF, \ourmethod and SVGP) are necessary for scalability (as highlighted by the next experiment), they remain inferior to the exact GPR model  -- which should be seen as the optimal baseline.

For VFF we have to resort to an additive model (A-VFF) in order to deal with the dimensionality of the data, as a vanilla VFF model can only be used in one or two dimensions. Following \citep[eq. 78]{hensman2017variational}, we assume a different function for each input dimension
\begin{equation}
\label{eq:vff-additive}
    f(x) = \sum_{d=1}^D f_d(x_d),\ \text{with}\ f_d \sim \GP\left(0, k_d\right),
\end{equation}
and approximate this process by a mean-field approximate posterior over the processes $q(f_1, \ldots, f_D) = \prod_{d} q(f_d)$, where each process is a SVGP $q(f_d) = \int p(f_d \given \vu_d) q(\vu_d) \calcd{\vu_d}$. We used $M=30$ frequencies per input dimension. As extra baseline we added an exact GPR model which makes the same additive assumption (A-GPR). As expected, not having to impose this structure improves performance; we see that \ourmethod beats A-VFF on every dataset.

\paragraph{Limitations}
Our current implementation of \ourmethod only supports datasets up to 8 dimensions (9 dimensions when the bias is concatenated). This is not caused by a theoretical limitation because our approach leads to a diagonal $\Kuu$ matrix in any dimension. The problem stems from the fact that there are no libraries available providing stable spherical harmonic implementations in high dimensions (needed for $k_{\vu}(\cdot)$). Our implementation based on \citet[Theorem~5.1]{dai2013} is stable up to 9 dimensions and future work will focus on scaling this up. Furthermore, \ourmethod does not solve  the curse of dimensionality for GP models but does drastically improve over the scaling of VFF.

\subsection{Large-Scale Regression on Airline Delay}

\begin{table*}[tb]
\centering
\resizebox{\textwidth}{!}{\begin{tabular}{lccccccccccc}
\toprule
& & \multicolumn{3}{c}{$N=10,000$} & \multicolumn{2}{c}{$N=100,000$} &
\multicolumn{2}{c}{$N = 1,000,000$}  & 
\multicolumn{3}{c}{$N = 5,929,413$}        \\
\cmidrule(lr){3-5}
\cmidrule(lr){6-7} 
\cmidrule(lr){8-9} 
\cmidrule(lr){10-12} 
Method & M 
& MSE & NLPD & Time
& MSE & NLPD
& MSE & NLPD
& MSE & NLPD & Time \\
\midrule
\ourmethod & 210 & 
$0.91 \pm 0.16$ &
$1.328 \pm 0.09$ &
$1.86 \pm 0.38$ &
$0.826 \pm 0.052$ &
$1.28 \pm 0.03$ &
$0.84 \pm 0.01$ &
$1.29 \pm 0.01$ &
$0.833 \pm 0.004$ &
$1.29 \pm 0.002$ &
$41.32 \pm 0.81$ \\
\ourmethod & 660 & 
$0.90 \pm 0.16$ &
\bm{$1.326 \pm 0.09$} &
$4.76 \pm 1.25$ &
$0.808 \pm 0.052$ &
\bm{$1.27 \pm 0.03$} &
$0.83 \pm 0.03$ &
\bm{$1.28 \pm 0.01$} &
$0.834 \pm 0.055$ &
\bm{$1.27 \pm 0.002$} &
$160.8 \pm 3.80$ \\
A-VFF & 30\small{/dim.} & 
 \bm{$0.89 \pm 0.15$} &
 $1.362 \pm 0.09$ &
 $6.78 \pm 0.85$ &
 $0.819 \pm 0.05$ &
 $1.32 \pm 0.03$ &
 $0.83 \pm 0.01$ &
 $1.33 \pm 0.03$ &
 $0.827 \pm 0.004$ &
 $1.32 \pm 0.007$ &
 $75.61 \pm 0.75$ \\
SVGP & 500 & 
 $0.90 \pm 0.16$ &
 $1.358 \pm 0.09$ &
 $836.54 \pm 0.78$ &
 \bm{$0.808 \pm 0.05$} &
 $1.31 \pm 0.03$ &
 \bm{$0.82 \pm 0.01$} &
 $1.32 \pm 0.002$ &
 \bm{$0.814 \pm 0.004$} &
 $1.31 \pm 0.002$ &
 $918.77 \pm 1.21$\\
\bottomrule
\end{tabular}

 }
\caption{Predictive mean squared errors (MSEs), negative log predictive densities (NLPDs) and wall-clock time in seconds with one standard deviation based on 10 random splits on the airline arrival delays experiment. Total dataset size is given by $N$ and in each split we randomly select 2/3 and 1/3 for training and testing.}
\label{tab:airline}
\end{table*}

This experiment illustrates three core capabilities of \ourmethod: 1) it can deal with large datasets and 2) it is computationally and time efficient 3) the model improves performance in terms of NLPD.

We use the 2008 U.S. airline delay dataset to asses these capabilities. The goal of this problem is to predict the amount of delay $y$ given eight characteristics $\vx$ of a flight, such as 
the age of the aircraft (number of years since deployment), route distance, airtime, etc. We follow the exact same experiment setup as \citet{hensman2017variational}\footnote{\url{https://github.com/jameshensman/VFF}} and evaluate the performance on 4 datasets of size 10,000, 100,000, 1,000,000, and 5,929,413 (complete dataset), created by subsampling the original one. For each dataset we use two thirds of the data for training and one third for testing. Every split is repeated 10 times and we report the mean and one standard deviation of the MSE and NLPD. For every run the outputs are normalized to be a centered unit Gaussian. The inputs are normalized to [0, 1] for VFF and SVGP. For \ourmethod we normalize the inputs so that each column falls within $[-v_d, v_d]$. The hyperparameter $v_d$ corresponds to the prior variance of the weights of an infinite-width fully-connected neural net layer (see \citet{cho2009kernel}). We can optimise for this weight-variance by back-propagation through $k_u(x)$ w.r.t. the ELBO. This is similar to the lengthscale hyperparameters of stationary kernels.

\Cref{tab:airline} shows the outcome of the experiment. The results for VFF and SVGP are from \citet{hensman2017variational}. We observe that \ourmethod improves on the other methods in terms of NLPD and is within error bars in terms of MSE. Given the variability in the data the GP models improve when more data is available during training.

Given the dimensionality of the dataset, a full-VFF model is completely infeasible. As an example, using just four frequencies per dimension would already lead to $M = 4^8 = 65,536$ inducing variables. So VFF has to resort to an additive model with a prior covariance structure given as a sum of Mat\'ern-3/2 kernels for each input dimension, as in \cref{eq:vff-additive}. Each of the functions $f_d$ is approximated using 30 frequencies.
We report two variants of \ourmethod: one using all spherical harmonics up to degree 3 ($M$=210) and another up to degree 4 ($M$=660). As expected, the more inducing variables, the better the fit.

We also report the wall clock time for the experiments (training and evaluation) for $N=10,000$ and $N=5,929,413$. All these experiments were ran on a single consumer-grade GPU (Nvidia GTX 1070). On the complete dataset of almost 6 million records, \ourmethod took $41\pm0.81$ seconds on average. A-VFF required $75.61\pm0.75$ seconds and the SVGP method needed approximately 15 minutes to fit and predict. This shows that \ourmethod is roughly two orders of magnitude faster than SVGP. A-VFF comes close to \ourmethod but has to impose additive structure to keep its computational advantage.

\subsection{SUSY Classification}

In the last experiment we tackle a large-scale classification problem. We are tasked with distinguishing between a signal process which produces super-symmetric (SUSY) particles and a background process which does not. The inputs consist of eight kinematic properties measured by the particle detectors in the accelerator. The dataset contains 5 million records of which we use the last 10\% for testing. We are interested in obtaining a calibrated classifier and measure the AuC of the ROC curve. 

For SVGP and \ourmethod we first used a subset of 20,000 points to train the variational and hyper-parameters of the model with L-BFGS. We then applied Adam to the whole dataset. A similar approach was used to fine-tune the NN baselines by \citet{baldi2014searching}.

\Cref{tab:susy} lists the performance of \ourmethod and compares it to a boosted decision tree (BDT), 5-layer neural network (NN), and a SVGP. We observe the competitive performance of \ourmethod, which is a single-layer GP method, compared to a 5-layer neural net with 300 hidden units per layer and extensive hyper-parameter optimisation \citep{baldi2014searching}. We also note the improvement over a SVGP with SE kernel.

\begin{table}[tb]
\centering
\scalebox{.9}{\begin{tabular}{@{}ll@{}}
\toprule
Method      & AuC             \\ \midrule
BDT         & $0.850  \pm 0.003$      \\
NN          & \bm{$0.867 \pm 0.002$}      \\
$\text{NN}_\text{dropout}$ & $0.856 \pm 0.001$    \\
SVGP (SE)        & $0.852 \pm 0.002$       \\
\ourmethod        & $0.859 \pm 0.001$       \\ \bottomrule
\end{tabular}}
\caption{Performance comparison for the SUSY benchmark.
The mean AuC is reported with one standard deviation, computed by training five models with different initialisations.  Larger is better. Results for BDT and NN are from \citet{baldi2014searching}.}
\vspace{-.3cm}
\label{tab:susy}
\end{table}

\section{Conclusion}

We introduced a framework for performing variational inference in Gaussian processes using spherical harmonics. Our general setup is closely related to VFF, and we inherit several of its advantages such as a considerable speed-up compared to classic sparse GP models, and having features with a global influence on the approximation. By projecting the data onto the hypersphere and using dedicated GP models on this manifold our approach succeeds where other sparse GPs methods fail. First, \ourmethod provides good scaling properties as the dimension of the input space increases.  Second, we showed that under some relatively weak hypothesis we are able to select the optimal features to include in the approximation. This is due to the intricate link between ``stationary'' covariances on the sphere and the Laplace-Beltrami operator. Third, the Mercer representation of the kernel means that the matrices to be inverted at training time are exactly diagonal -- resulting in a very cheap-to-compute sparse approximate GP.

We showed on a wide range of regression and classification problems that our method performs at or close to the state of the art while being extremely fast. The good predictive performance may be inherited from the connection between infinitely wide neural network and the way we map the predictions on the sphere back to the original space. Future work will explore this hypothesis.

\subsection*{Acknowledgements}
Thanks to Fergus Simpson for pointing us in the direction of spherical harmonics in the early days of this work. Also many thanks to Arno Solin for sharing his implementation of the Mat\'ern kernels' power spectrum.

\bibliography{ref}
\bibliographystyle{icml2020}

\clearpage
\raggedbottom

\includepdf[pages=-]{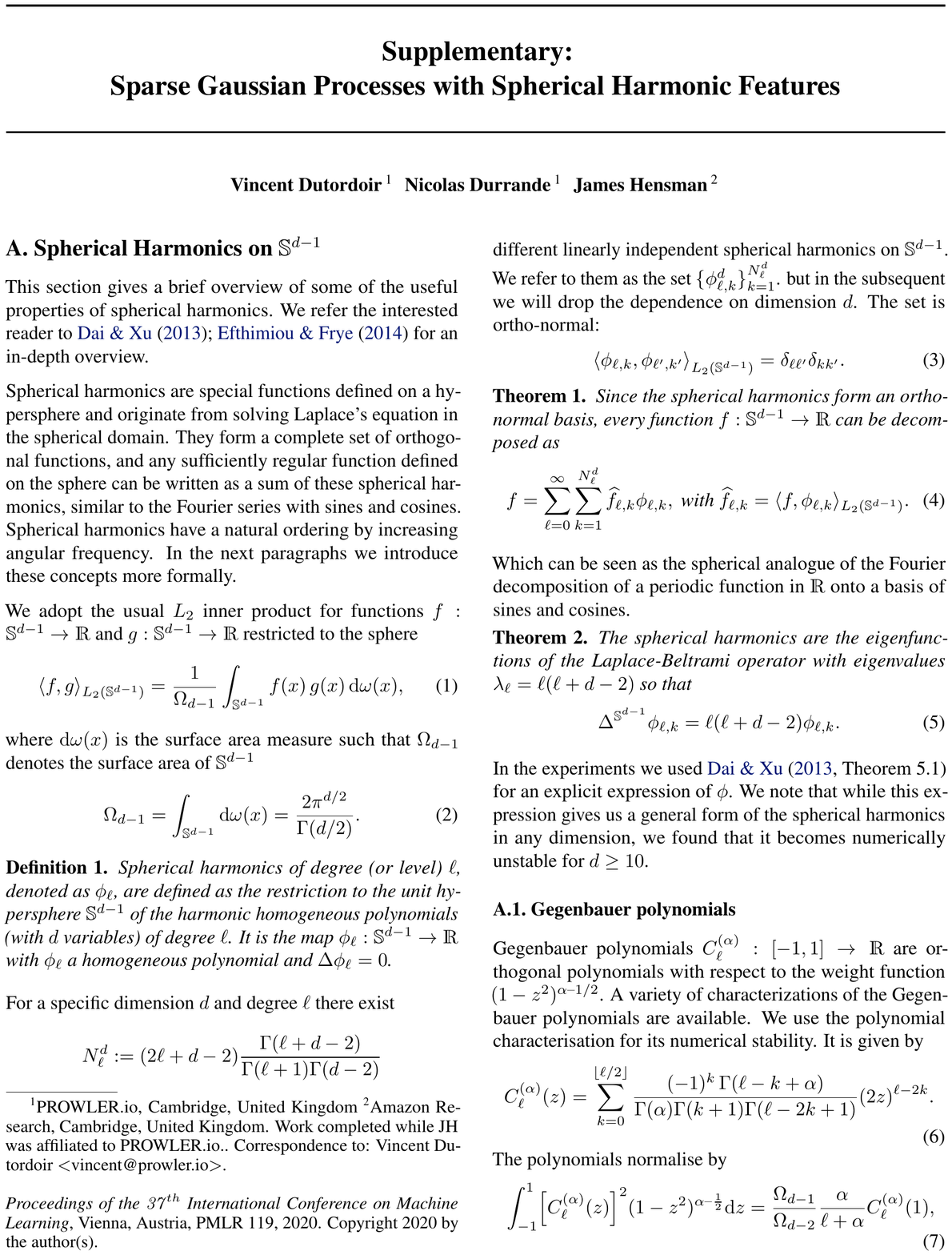}

\makeatletter\@input{suppl_aux.tex}\makeatother

\end{document}

\twocolumn[
\icmltitle{Supplementary:\\Sparse Gaussian Processes with Spherical Harmonic Features} 

\begin{icmlauthorlist}
\icmlauthor{Vincent Dutordoir}{pio}
\icmlauthor{Nicolas Durrande}{pio}
\icmlauthor{James Hensman}{amazon}
\end{icmlauthorlist}

\icmlaffiliation{pio}{PROWLER.io, Cambridge, United Kingdom}
\icmlaffiliation{amazon}{Amazon Research, Cambridge, United Kingdom. Work completed while JH was affiliated to PROWLER.io.}

\icmlcorrespondingauthor{Vincent Dutordoir}{vincent@prowler.io}

\icmlkeywords{Gaussian processes, spherical harmonics, variational inference, probabilistic modelling}

\vskip 0.3in
]
\printAffiliationsAndNotice{}

% \onecolumn

\section{Spherical Harmonics on \dsphere}

This section gives a brief overview of some of the useful properties of spherical harmonics. We refer the interested reader to \citet{dai2013, frye2014} for an in-depth overview.

Spherical harmonics are special functions defined on a hypersphere and originate from solving Laplace's equation in the spherical domain. They form a complete set of orthogonal functions, and any sufficiently regular function defined on the sphere can be written as a sum of these spherical harmonics, similar to the Fourier series with sines and cosines. Spherical harmonics have a natural ordering by increasing angular frequency. In the next paragraphs we introduce these concepts more formally.

We adopt the usual $L_2$ inner product for functions $f: \dsphere \rightarrow \Reals$ and $g: \dsphere \rightarrow \Reals$ restricted to the sphere 
\begin{equation}
     \langle f, g\rangle_{L_{2}(\dsphere)} = \frac{1}{\darea} \int_{\dsphere} f(x)\,g(x) \, \calcd{\omega(x)},
\end{equation}
where $\calcd{\omega(x)}$ is the surface area measure such that $\darea$ denotes the surface area of $\dsphere$ 
\begin{equation}
    \darea = \int_{\dsphere} \calcd{\omega(x)} = \frac{2 \pi ^ {d/2}}{\Gamma(d/2)}.
\end{equation}

\begin{definition}
Spherical harmonics of degree (or level) $\ell$, denoted as $\sh_\ell$, are defined as the restriction to the unit hypersphere $\dsphere$ of the harmonic homogeneous polynomials (with $d$ variables) of degree $\ell$. It is the map $\sh_\ell : \dsphere \rightarrow \Reals$ with $\sh_\ell$ a homogeneous polynomial and $\Delta \sh_\ell = 0$. 
\end{definition}

For a specific dimension $d$ and degree $\ell$ there exist 
$$
\dnumharmonicsforlevel := (2 \ell + d - 2) \frac{\Gamma(\ell + d -2)}{\Gamma(\ell + 1)\Gamma(d - 2)}
$$
different linearly independent spherical harmonics on $\dsphere$. We refer to them as the set $\{ \sh_{\ell, k}^d \}_{k=1}^{\dnumharmonicsforlevel}$. but in the subsequent we will drop the dependence on dimension $d$. The set is ortho-normal:
\begin{equation}
    \left\langle \sh_{\ell, k}, \sh_{\ell', k'}\right\rangle_{L_2(\dsphere)} 
    % = \frac{1}{\darea} \int_{\dsphere} \sh_\ell(x)\,\sh_{\ell'}(x) \, \calcd{\omega(x)}
    =\delta_{\ell \ell'} \delta_{k k'}.
\end{equation}

\begin{theorem}
Since the spherical harmonics form an ortho-normal basis, every function $f: \dsphere \rightarrow \Reals$ can be decomposed as
\begin{equation}
    f = \sum_{\ell=0}^{\infty} \sum_{k=1}^{\dnumharmonicsforlevel} \widehat{f}_{\ell, k} \sh_{\ell, k},\ \text{with}\ \widehat{f}_{\ell, k} = \langle f,  \sh_{\ell, k} \rangle_{L_2(\dsphere)}.
\end{equation}
\end{theorem}

Which can be seen as the spherical analogue of the Fourier decomposition of a periodic function in $\Reals$ onto a basis of sines and cosines.

\begin{theorem}
\label{appendix:theorem:eigenvalues}
The spherical harmonics are the eigenfunctions of the Laplace-Beltrami operator with eigenvalues $\lambda_{\ell} = \ell (\ell + d - 2)$ so that
\begin{equation}
    \Delta^{\dsphere} \phi_{\ell, k} = \ell (\ell + d - 2) \phi_{\ell, k}.
\end{equation}
\end{theorem}

In the experiments we used \citet[Theorem~5.1]{dai2013} for an explicit expression of $\phi$. We note that while this expression gives us a general form of the spherical harmonics in any dimension, we found that it becomes numerically unstable for $d \ge 10$. 

\subsection{Gegenbauer polynomials}

Gegenbauer polynomials $C_\ell^{(\alpha)}: [-1, 1] \rightarrow \Reals$ are orthogonal polynomials with respect to the weight function $(1 - z^2)^{\alpha–1/2}$.
A variety of characterizations of the Gegenbauer polynomials are available. We use the polynomial characterisation for its numerical stability. It is given by
\begin{equation}
\label{appendix:eq:gegenbauer}
    C_{\ell}^{({\alpha})}(z)=\sum _{{k=0}}^{{\lfloor \ell/2\rfloor }}{\frac  {(-1)^{k}\, \Gamma (\ell-k+\alpha )}{\Gamma (\alpha )\Gamma({k+1})\Gamma{(\ell-2k + 1)}}}(2z)^{{\ell-2k}}.
\end{equation}
The polynomials normalise by
\begin{equation}
    \int_{-1}^{1}  \left[C_{\ell}^{({\alpha})}(z)\right]^2  (1 - z^2)^{\alpha–\frac{1}{2}} \calcd{z} = \frac{\Omega_{d-1}}{\Omega_{d-2}} \frac{\alpha}{\ell + \alpha} C_{\ell}^{({\alpha})}(1),
\end{equation}
with $C_{\ell}^{({\alpha})}(1) = \frac{\Gamma(2\alpha + \ell)}{\Gamma(\alpha)\,\ell!}$.

There exists a close relationship between Gegenbauer polynomials (also known as \emph{generalized Legendre polynomials}) and spherical harmonics, as we will show in the next theorems. 

\begin{theorem}[Addition]
\label{appendix:theorem:addition}
Between the spherical harmonics of degree $\ell$ in dimension $d$ and the Gegenbauer polynomials of degree $\ell$ there exists the relation
% \begin{equation}
%     \sum_{k=1}^{\dnumharmonicsforlevel} \sh_{\ell, k}(\vx) \sh_{\ell, k}(\vx') = {\dnumharmonicsforlevel} C_\ell^{\frac{d-2}{2}}(\vx\transpose\vx').
% \end{equation}
\begin{equation}
    \sum_{k=1}^{\dnumharmonicsforlevel} \sh_{\ell, k}(\vx) \sh_{\ell, k}(\vx') = \frac{\ell + \alpha}{\alpha}\,
    C_\ell^{(\alpha)}(\vx\transpose\vx'),
\end{equation}
with $\alpha = \frac{d-2}{2}$.
\end{theorem}

As a illustrative example, this property is analogues to the trigonometric addition formula: $\sin(x)\sin(x') + \cos(x)\cos(x') = \cos(x - x')$.

\begin{theorem}[Funk-Hecke]
\label{appendix:theorem:funk}
Let $s(\cdot)$ be an integrable function such that $\int_{-1}^1 \| s(t)\| (1 - t^2)^{(d-3)/2} \calcd{t}$ is finite and $d \ge 2$. Then for every $\sh_{\ell,k}$ 
\begin{equation}
    \frac{1}{\darea} \int_{\dsphere} s(\vx\transpose \vx')\,\sh_{\ell, k}(\vx')\, \calcd{\omega(\vx')} = \widehat{a}_{\ell}\,\sh_{\ell,k}(\vx),
\end{equation}
where $\widehat{a}_{\ell}$ is a constant defined by
\begin{equation}
    \widehat{a}_{\ell}  = 
    % \frac{\Omega_{d-2}}{\Omega_{d-1}} 
    \frac{\omega_{d}}{C_\ell^{(\alpha)}(1)} \int_{-1}^1 s(t)\,C_\ell^{(\alpha)}(t)\,(1 - t^2)^{\frac{d-3}{2}} \calcd{t},
\end{equation}
with $\alpha = \frac{d-2}{2}$ and $\omega_d = \frac{\Omega_{d-2}}{\Omega_{d-1}}$.
\end{theorem}

Funk-Hecke simplifies a $(d-1)$-variate surface integral on $\dsphere$ to a one-dimensional integral over $[-1, 1]$. This theorem gives us a practical way of computing the Fourier coefficients for any zonal kernel. In \cref{sec:appendix:arc-cosine}, we use it to compute the coefficients of the arc-cosine kernel. Notice how the Fourier coefficients $\widehat{a}_{\ell}$ only depend on the level $\ell$ (or degree) of the spherical harmonic and not the orientation (denoted by the $k$ index).

\section{Zonal kernels}
\label{appendix:sec:zonal-kernels}

\subsection{Mercer's decomposition}
Zonal kernels can be seen as the spherical counterpart of stationary kernels. Stationary kernels are a function of $\vx - \vx'$ and are thus invariant to translations in the input space \citep{rasmussen2006}. Zonal kernels (defined on $\dsphere \times \dsphere$) are a function of $\vx^\top \vx'$ and are thus invariant to rotations.

The spherical harmonics are the eigenfunctions of the Laplace-Beltrami operator \cite{dai2013, frye2014}. In the main paper we show the commutativity of the Laplace-Beltrami operator and the kernel operator of zonal kernels. This means that the spherical harmonics are also the eigenfunctions of zonal kernels, as commuting operators share the same eigenfunctions. 

Mercer's theorem allows us to express the kernel in terms of its eigenvalues and eigenfunctions.
\begin{theorem}[Mercer representation]
\label{appendix:theorem:mercer}
Any zonal kernel $k$ on the hypersphere can be decomposed as
\begin{equation}
    k(\vx, \vx') = \sum_{\ell=0}^{\infty} \sum_{k=1}^{\dnumharmonicsforlevel} \widehat{a}_{\ell, k} \sh_{\ell, k}(\vx) \sh_{\ell, k}(\vx'),
\end{equation}
where $\vx,\vx' \in \dsphere$ and $\widehat{a}_{\ell}$ are the positive Fourier coefficients, $\sh_{\ell,k}$ denote the elements of the spherical harmonic basis in $\dsphere$, and $N_\ell^d$ corresponds to the number of spherical harmonics for a given level $\ell$.
\end{theorem}

For zonal kernels the Fourier coefficients within a level are equal: $\widehat{a}_{\ell, k} = \widehat{a}_{\ell}$ for $1 \le k \le N_\ell^d$. This allows us to simplify the Mercer decomposition of a zonal kernel using \cref{appendix:theorem:addition} to 
\begin{equation}
    k(\vx, \vx') = \sum_{\ell=0}^{\infty} \widehat{a}_{\ell}
    \frac{\ell + \alpha}{\alpha}\,
    C_\ell^{(\alpha)}(\vx\transpose\vx'),
\end{equation}
with $\alpha = \frac{d-2}{2}$.

\subsection{RKHS}
Given the Mercer representation of a zonal kernel, its RKHS can be characterised by
\begin{equation*}
    \rkhs = \left\{
    g = 
    \sum_{\ell=0}^\infty \sum_{k=1}^{\dnumharmonicsforlevel} \widehat{g}_{\ell,k} \sh_{\ell, k} :
    \sum_{\ell=0}^\infty \sum_{k=1}^{\dnumharmonicsforlevel} \frac{|\widehat{g}_{\ell,k}|^2}{\widehat{a}_{\ell}} < \infty
    \right\}
\end{equation*}
with a reproducing inner product between two functions $g(\vx) = \sum_{\ell, k} \widehat{g}_{\ell, k} \sh_{\ell, k}(\vx)$ and $h(\vx) = \sum_{\ell, k} \widehat{h}_{\ell, k} \sh_{\ell, k}(\vx)$ defined as
\begin{equation}
\label{appendix:eq:inner-product}
    \langle g, h \rangle_{\rkhs} = 
        \sum_{\ell=0}^\infty 
        \sum_{k=1}^{\dnumharmonicsforlevel}
            \frac{\widehat{g}_{\ell, k} \widehat{h}_{\ell, k}}%
                 {\widehat{a}_{\ell}}.
\end{equation}
\begin{proof}{(Reproducing property).}
\label{appendix:proof:reproducing}
The Fourier coefficients for $k(\vx, \cdot): \dsphere \rightarrow \Reals$ and $f: \dsphere \rightarrow \Reals$ are $\widehat{a}_{\ell, k} \sh_{\ell, k}(\vx)$ and $\widehat{f}_{\ell, k}$, respectively.
Substituting these coefficients in \cref{appendix:eq:inner-product} gives:
\begin{align}
    \langle k(\vx, \cdot), f \rangle_{\rkhs} 
    & = \sum_{\ell=0}^\infty \sum_{k=1}^{\dnumharmonicsforlevel} 
    \frac{\widehat{a}_{\ell} \sh_{\ell, k}(\vx) \widehat{f}_{\ell,k}}{\widehat{a}_{\ell}}\\
    &= \sum_{\ell=0}^\infty \sum_{k=1}^{\dnumharmonicsforlevel} 
    \widehat{f}_{\ell,k} \sh_{\ell, k}(\vx) = f(\vx)
\end{align}
which proofs the reproducing property.
\end{proof}

In the next sections we address the computation of the Fourier coefficients (eigenvalues) of the kernels. In \cref{sec:appendix:arc-cosine} for the Arc-Cosine kernel and in \cref{appendix:sec:materns} for the Mat\'ern family.
\subsection{Fourier coefficients for the Arc-Cosine kernel}
\label{sec:appendix:arc-cosine}

The Fourier coefficients are computed using \cref{appendix:theorem:funk}, where the shape function of the Arc-Cosine kernel of the first order \citep{cho2009kernel} is given by:
\begin{equation}
    s(x) = \sin x + (\pi - x) \cos x.
\end{equation}
Notice that we expressed the shape function as a function of the angle between the two inputs $s: [0, \pi] \mapsto \Reals$, rather than the great-circle distance, as it simplifies the subsequent computations.

Using a change of variables we also rewrite \cref{appendix:theorem:funk}
\begin{equation}
    \widehat{a}_{\ell}
    =  c_{d, \ell}  \int_{0}^\pi\!\!s(x)\,C_\ell^{\frac{d-2}{2}}(\cos x) \sin^{d-2} x\,\calcd{x},
\end{equation}
with $c_{d, \ell} = \frac{\omega_{d}}{C_\ell^{(\alpha)}(1)}$.
This one dimensional integral can be solved in closed-form for any setting of $d$ and $\ell$. Filling in the definition for the Gegenbauer polynomial (\cref{appendix:eq:gegenbauer}), we observe that we need a general solution of the integral $\int_0^\pi \left[ \sin(x) + (\pi -x) \cos(x) \right] \cos^n(x) \sin^m(x) \diff x$ for $n, m \in \mathbb{N}$.

The first term can be computed with the well-known result:
% https://math.stackexchange.com/questions/2833731/reduction-formula-for-integral-sinm-x-cosn-x-with-limits-0-to-pi-2
\begin{gather}
    \int_0^{\pi} \sin^n(x) \cos^m(x) \diff x \\ =
    \begin{cases}
        0                                   & \text{if}\ m\ \text{odd}\\
        \frac{(n-1)!!~(m-1)!!}{(n+m)!!} \pi & \text{if}\ m\ \text{even and }n\ \text{odd},\\
        \frac{(n-1)!!~(m-1)!!}{(n+m)!!} 2   & \text{if}\ n,m\ \text{even}.
    \end{cases}
\end{gather}

The second term is harder:
\begin{equation}
    I = \int_0^{\pi} (\pi - x) \sin^n(x) \cos^m(x) \diff x
\end{equation}
which we solved using integration by parts with $u = \pi - x$ and $dv = \sin^n(x) \cos^m(x) dx$, so that 
\begin{equation}
    I = u(0) v(0) - u(\pi) v(\pi) + \int_0^{\pi} v(x') \diff x',
\end{equation}
where $v(x') = \int_0^{x'} \sin^n(x) \cos^m(x) \diff x$. This gives $v(0) = 0$ and $u(0) = 0$, simplifying $I = \int_0^\pi v(x') \diff x'$.
% \begin{equation}
%     I = \int_0^{\pi} \int_0^x \sin^n(x') \cos^m(x') \diff x' \diff x,
% \end{equation}
We first focus on $v(x')$:
for \underline{$n$ odd}, there exists a $n' \in \mathbb{N}$ so that $n = 2n' + 1$, resulting
\begin{align}
    v(x') &= \int_0^{x'} \sin^{2n'}(x) \cos^m(x) \sin(x) \diff x \\
          &= -\int_0^{\cos(x')} (1 - u^2)^{n'} u^m \diff u
\end{align}
Where we used $\sin^2(x) + \cos^2(x) = 1$ and the substitution $u = \cos(x) \implies \diff u = - \sin(x) \diff x$.
Using the binomial expansion, we get
\begin{align}
    v(x') &= -\int_0^{\cos(x')} \sum_{i=0}^{n'} \binom{k}{i} (-u^2)^i u^m \diff u \\
    &= \sum_{i=0}^{n'} (-1)^{i+1} \binom{k}{i} \frac{\cos(x')^{2i+m+1} - 1}{2i+m+1}.
\end{align}

Similarly, for \underline{$m$ odd}, we have $m=2m' + 1$ and use the substitution $u = \sin(x)$, to get
\begin{equation}
    v(x') = \sum_{i=0}^{m'} (-1)^{i} \binom{k}{i} \frac{\sin(x')^{2i+n+1}}{2i+n+1}.
\end{equation}

For \underline{$n$ and $m$ even}, we have $n' = n/2$ and $m' = m/2$, we use double-angle identities to get
\begin{equation}
    v(x') = \int_0^{x'} \left(\frac{1 - \cos(2x)}{2}\right)^{n'} \left(\frac{1 + \cos(2x)}{2}\right)^{m'} \diff x
\end{equation}
Making use of the binomial expansion twice, we get
\begin{equation}
    v(x') = 2^{-(n' + m')} \sum_{i,j=0}^{n', m'} (-1)^{i} \binom{n'}{i} \binom{m'}{j} 
    \int_0^{x'} \cos(2x)^{i+j} \diff x.
\end{equation}

Returning back to the original problem $I = \int_0^\pi v(x') \diff x'$. Depending on the parity of $n$ and $m$ we need to evaluate:
% $\int_0^\pi \cos(x')^p \diff x'$ ($n$ odd),
% $\int_0^\pi \sin(x')^p \diff x'$ ($m$ odd) and
% $\int_0^\pi \int_0^{x'} \cos(2x)^p \diff x \diff x'$ ($n$ and $m$ even), which are given by
\begin{equation}
    \int_0^\pi \cos(x')^p \diff x' = 
    \begin{cases}
        \frac{(p-1)!!}{p!!} \pi   & \text{if}\ p\ \text{even} \\
        0                         & \text{if}\ p\ \text{odd}
    \end{cases}
\end{equation}
and
\begin{equation}
    \int_0^\pi \sin(x')^p \diff x' = 
    \begin{cases}
        \frac{(p-1)!!}{p!!} \pi   & \text{if}\ p\ \text{even} \\
        \frac{(p-1)!!}{p!!} 2   & \text{if}\ p\ \text{odd}.
    \end{cases}
\end{equation}
For $m$ and $n$ even we need to solve the double integral
\begin{equation}
    \int_0^\pi \int_0^{x'} \cos(2x)^p \diff x \diff x' = 
    \begin{cases}
        \frac{(p-1)!!}{p!!} \frac{\pi^2}{2}   & \text{if}\ p\ \text{even} \\
        0   & \text{if}\ p\ \text{odd}.
    \end{cases}
\end{equation}

Combining these results gives us the solution for the integral $\int_0^\pi s(x) \cos^n(x) \sin^m(x) \diff x$ for any $n, m \in \Naturals$, which is necessary to compute $\widehat{a}_\ell$ for the arc-cosine kernel.

\subsection{Fourier coefficients for the Mat\'ern family kernels}
\label{appendix:sec:materns}
The Mat\'ern covariance between two points $x, x'$ separated by $r=x-x'$ distance units is given by \citep{rasmussen2006}:
\begin{equation}
    k_\nu(r)=\sigma ^{2}{\frac {2^{1-\nu }}{\Gamma (\nu )}}{\Bigg (}{\sqrt {2\nu }}{\frac {r}{\rho }}{\Bigg )}^{\nu }K_{\nu }{\Bigg (}{\sqrt {2\nu }}{\frac {r}{\rho }}{\Bigg )},
\end{equation}
where $\Gamma$  is the gamma function, $K_{\nu}$ is the modified Bessel function of the second kind, and $\rho$ (lengthscale) and $\nu$ (differentiability) are non-negative parameters of the covariance. The covariance has a spectral density defined on $\Reals^d$
\begin{equation}
    \displaystyle S(\omega)={\frac {2^{d}\pi ^{\frac {d}{2}}\Gamma (\nu +{\frac {d}{2}})(2\nu )^{\nu }}{\Gamma (\nu )\rho ^{2\nu }}}\left({\frac {2\nu }{\rho ^{2}}}+4\pi ^{2}\omega^{2}\right)^{-\left(\nu +{\frac {d}{2}}\right)}.
\end{equation}

A key result from \citet[eq. 20]{solin2014hilbert} is to show that the coefficients $\widehat{a}_{\ell, k}$ have a simple expression that depends on the kernel spectral density $S$ and the eigenvalues of the Laplace-Beltrami operator (\cref{appendix:theorem:eigenvalues}). For GPs on $\dsphere$ the coefficients boil down to 
\begin{equation}
    \widehat{a}_{\ell, k } =  S\left(\sqrt{\ell (\ell + d - 2)}\right).
\end{equation}

\raggedbottom

\bibliography{ref}
\bibliographystyle{icml2020}